\documentclass[11pt]{article}
\usepackage[top=1in, bottom=1in, left=1in, right=1in]{geometry}

\usepackage{times}
\usepackage{helvet}
\usepackage{courier}
\usepackage{hyperref}
\usepackage{url}
\usepackage{amsmath}
\usepackage{amssymb}
\usepackage{amsthm}
\usepackage{enumitem}
\usepackage{graphicx}
\usepackage{pdfpages}
\usepackage{bm}

\newif\ifcomment
\commenttrue      

\ifcomment
\newcommand{\qirong}[1]{{\color{blue}{\bf\sf [Qirong: #1]}}} 

\newcommand{\daiwei}[1]{{\color{purple}{\bf\sf [Daiwei: #1]}}}
\newcommand{\jinliang}[1]{{\color{red}{\bf\sf [Jinliang: #1]}}}

\usepackage{soul}	
\newcommand{\ericw}[1]{{\color{blue}{#1}}}
\newcommand{\ericx}[1]{{\color{magenta}{\sf \bf [Eric: #1]}}}
\newcommand{\eric}[1]{\marginpar{\color{blue}\tiny{Eric: #1}}}

\else

\newcommand{\qirong}[1]{{\color{blue}{}}}
\newcommand{\abhi}[1]{{\color{magenta}{}}}
\newcommand{\daiwei}[1]{{\color{purple}{}}}
\newcommand{\jinliang}[1]{{\color{green}{}}}
\newcommand{\ericw}[1]{{\color{green}{}}}
\newcommand{\ericx}[1]{{\color{green}{}}}
\newcommand{\eric}[1]{{\color{green}{}}}
\fi

\newtheorem{theorem}{Theorem}
\newtheorem{assumption}{Assumption}

\newtheorem{lemma}[theorem]{Lemma}

\newcommand{\bx} {\textbf{x}}
\newcommand{\hbu} {\hat{\textbf{u}}}
\newcommand{\hbx} {\hat{\textbf{x}}}
\newcommand{\bbx} {\breve{\textbf{x}}}
\newcommand{\bbg} {\breve{\bm{g}}}
\newcommand{\mE} {\mathbb{E}}
\newcommand{\bgamma} {\bm{\gamma}}
\newcommand{\bdelta} {\bm{\delta}}

\newcommand{\tbx} {\tilde{\textbf{x}}}

\newcommand{\bu} {\textbf{u}}

\newcommand{\bigo} {\mathcal{O}}
\newcommand{\order}[1]{\mathcal{O}(#1)}
\def\Var{{\rm Var}\,}

\begin{document}
%
\title{High-Performance Distributed ML at Scale through Parameter Server Consistency Models}
\author{Wei Dai, Abhimanu Kumar, Jinliang Wei, Qirong Ho*, Garth Gibson and Eric P. Xing
\\
School of Computer Science, Carnegie Mellon University
\\
*Institute for Infocomm Research, A*STAR
\\
\texttt{wdai,abhimank,jinlianw,garth,epxing@cs.cmu.edu, hoqirong@gmail.com}
}
\date{}

\maketitle
\begin{abstract}
\begin{quote}
As Machine Learning (ML) applications increase in data size and model complexity, practitioners
turn to distributed clusters to satisfy the increased computational and memory demands.
Unfortunately, effective use of clusters for ML requires considerable expertise in writing distributed code,
while highly-abstracted frameworks like Hadoop have not, in practice,
approached the performance seen in specialized ML implementations. The recent
Parameter Server (PS) paradigm is a middle ground between these extremes,
allowing easy conversion of single-machine parallel ML applications into distributed
ones, while maintaining high throughput through relaxed ``consistency models" that allow inconsistent
parameter reads. However, due to insufficient theoretical study, it is not clear which of
these consistency models can really ensure correct ML algorithm output; at the same time,
there remain many theoretically-motivated but undiscovered opportunities to maximize computational throughput.
Motivated by this challenge, we study both the theoretical guarantees and empirical behavior of
{\it iterative-convergent ML algorithms} in existing PS consistency models. We then
use the gleaned insights to improve a consistency model using an ``eager" PS communication mechanism,
and implement it as a new PS system that enables ML algorithms to reach their solution more
quickly.
\end{quote}
\end{abstract}
\vspace{-0.3cm}
\section{Introduction}

The surging data volumes generated by internet activity and scientific research~~\cite{dean2012large}
put tremendous pressure on Machine Learning (ML) methods to scale beyond the computation and memory
of a single machine. On one hand, very large data sizes (Big Data) require too
much time for complex ML models to process on a single machine~\cite{Ahmed:2012:SIL:2124295.2124312,ho2013more,cipar2013,cui_atc_14,muli_osdi14},
which necessitates distributed-parallel computation over an entire cluster
of machines. A typical solution to this problem is \emph{data paralllelism}, in which the data is partitioned
and distributed across different machines, which train the (shared) ML model using their local data. In order to
share the model across machines, practitioners have recently turned to a "Parameter server" (PS) paradigm~\cite{Ahmed:2012:SIL:2124295.2124312,ho2013more,cipar2013,cui_atc_14,muli_osdi14}.

Many general-purpose Parameter Server (PS) systems~\cite{ho2013more,cipar2013,cui_atc_14} of ML computation
provide a Distributed Shared Memory (DSM) solution to the Big Data and Big Model issues. DSM allows ML programmers to treat the entire cluster as a single memory pool, where every machine can read/write to any model parameter via a simple programming interface; this greatly simplifies the implementation of distributed
ML programs, because programmers may treat a cluster like a ``supercomputer" that can run thousands of computational threads, without worrying about low-level communication between machines. It should be noted that not all PS systems provide a DSM interface; some espouse an arguably less-convenient push/pull interface that requires users to explicitly decide which parts of the ML model need to be communicated~\cite{muli_osdi14}.


At the same time, the iterative-convergent nature of ML programs presents unique opportunities and challenges that
do not manifest in traditional database applications: for example, ML programs lend themselves well to stochastic subsampling or randomized algorithms, but at the same time exhibit complex dependencies or correlations between parameters that can make parallelization difficult~\cite{Bradley+al:icml11parlasso,lee2014strads}.
Recent works~\cite{niu2011hogwild,ho2013more,fugue} have introduced relaxed consistency models to trade off between parameter read accuracy and read throughput, and show promising speedups over fully-consistent models; their success is underpinned by the error-tolerant nature of ML, that ``plays nicely'' with relaxed synchronization guarantees --- and in turn, relaxed synchronization allows system designers to achieve higher read throughput, compared to fully-consistent models.

However, we still possess limited understanding of (1) how relaxed consistency affects ML algorithmconvergence rate and stability, and (2) what opportunities still exist for improving the performance of boththe ML algorithm (how much progress it makes per iteration), and the throughput of the PS system (howmany  ML  algorithm  iterations  can  be  executed  per  second).  Recent works on PS have only focused on system optimizations in PS using various heuristics like async relaxation~\cite{adam} and uneven updates propagation based on parameter values~\cite{muli_osdi14}. Our work instead  starts from an ML-theoretic standpoint and provides  principled  insights  to improve  PS design. Concretely, we examine the theoretical and empirical behavior of PS consistency models from new angles, such as the distribution of stale reads and the impact of staleness on solution stability. We then apply the learnt insights to design more efficient consistency models and PS system that outperform previous work.

To study these issues, we formulate a new Value-Bounded Asynchronous Parallel (VAP) model, and show that it provides an ideal, gold-standard target in terms of theoretical behavior (high progress per ML algorithm iteration). However, VAP, of which the basic idea or principle is attempted in~\cite{li2013parameter}, can be problematic because bounding the value of in-transit updates amounts to tight synchronization.
We propose Eager Stale Synchronous Parallel (ESSP), a variant of Stale Synchronous Parallel (SSP, a bounded-iteration model that is fundamentally different from VAP) introduced in~\cite{ho2013more}, and formally and empirically show that ESSP is a practical and easily realizable scheme for parallelization.
Specifically, we develop new variance bounds for both ESSP and VAP, and show that ESSP attains the same guarantees as VAP. These variance bounds provide a deeper characterization of convergence (particularly solution stability) under SSP and VAP, unlike existing PS theory that is focused only on expectation bounds~\cite{ho2013more}. We develop an efficient implementation of ESSP and shows that it outperforms SSP in convergence (both real time and per iteration) by reducing the average staleness, consistent with our theoretical results. 



\vspace{-0.3cm}
\section{Consistency Models for Parameter Servers}
\vspace{-0.1cm}
\label{sec:consistency}


A key idea for large-scale distributed ML is to carefully trade off parameter consistency for increased parameter read throughput (and thus faster algorithm execution), in a manner that guarantees the final output of an ML algorithm is still ``correct" (meaning that it has reached a locally-optimal answer). This is possible because ML algorithms are {\it iterative-convergent} and {\it error-tolerant}: 
ML algorithms will converge to a local optimum even when there are
errors in the algorithmic procedure itself (such as stochasticity in randomized methods).



In a distributed-parallel environment,
multiple workers must simultaneously generate updates to shared global parameters. Hence, enforcing strong consistency (parameters updates are immediately reflected) quickly leads to frequent, time-consuming synchronization and thus very limited speed up from parallelization.
One must therefore define a relaxed consistency model that enables low-synchronization parallelism while closely approximating the strong consistency of sequential execution. The insight is that, to an iterative-convergent ML algorithm, inconsistent parameter reads have essentially the same effect as errors due to the algorithmic procedure --- implying that convergence to local optima can still happen even under inconsistent reads, {\it provided the degree of inconsistency is carefully controlled}. We now explain two possible PS consistency models, and the trade-offs they introduce.

\vspace{-0.3cm}
\subsection{The ideal but inefficient Value-bounded Asynchronous Parallel (VAP) model}
\vspace{-0.1cm}
We first introduce Value-bounded Asynchronous Parallel (VAP), an ideal model that directly approximates strong consistency (e.g. in the sequential setting) by bounding the difference in {\it magnitude} between the strongly consistent view of values (i.e. values under the single-thread model) and the actual parameter views on the workers. Formally, let ${\bf x}$ represent all model parameters, and assume that each worker in the ML algorithm produces additive updates (${\bf x}\leftarrow {\bf x}+ {\bf u}$, where ${\bf u}$ is the update)\footnote{This is common in algorithms such as gradient descent (${\bf u}$ being the gradient) and sampling methods.}. Given $P$ workers, we say that an update ${\bf u}$ is {\it in transit} if ${\bf u}$ has been seen by $P-1$ or fewer workers --- in other words, it is yet visible by all workers. Update ${\bf u}$ is no longer in transit once seen by all workers. The VAP requires the following condition:

\paragraph{VAP condition:}
\vspace{-0.3cm}
Let ${\bf u}_{p,i}$ be the updates from worker $p$ that are in transit, and ${\bf u}_p := \sum_i {\bf u}_{p,i}$. VAP requires that, whenever any worker performs a computation involving the model variables ${\bf x}$, the condition $ ||{\bf u}_p ||_{\infty} \le v_{thr}$ holds for a specified (time-varying) value bound parameter $v_{thr}$. In other words, the aggregated in-transit updates from all workers cannot be too large.

To analyze VAP, we must identify algorithmic properties common to ML algorithms. Broadly speaking, most ML algorithms are either {\it optimization-based} or {\it sampling-based}. Within the former, many Big Data ML algorithms are stochastic gradient descent-based (SGD), because SGD allows each worker to operate on its own data partition (i.e. no need to transfer data between workers), while the algorithm parameters are globally shared (hence a PS is necessary). SGD's popularity makes it a good choice for grounding our analysis ---
in later section, we show that VAP approximates strong consistency well in these senses: (1) SGD with VAP errors converges {\it in expectation} to an
optimum; (2) the parameter {\it variance} decreases in successive iterations, guaranteeing the quality and stability of the final result.


While being theoretically attractive, the VAP condition is too strong to be implemented efficiently in practice: before any worker can perform computation on ${\bf x}$, it must ensure that the in-transit updates from all other workers sum to at most $v_{thr}$ component-wise due to the max-norm. This poses a chicken-and-egg conundrum: for a worker to ensure the VAP condition holds, it needs to know the updates from all other workers --- which, in general, requires the same amount of communication as strong consistency, defeating the purpose of VAP. While it may be possible to relax the VAP condition for specific problem structures, in general, value-bounds are difficult to achieve for a generic PS.

\vspace{-0.3cm}
\subsection{Eager Stale Synchronous Parallel (ESSP)}
\vspace{-0.1cm}
In order to design a consistency model that is practically efficient while providing proper correctness guarantees, we consider an iteration-based consistency model called Stale Synchronous Parallel (SSP)~\cite{ho2013more}, that lends itself to an efficient PS implementation. At a high level, SSP imposes bounds on {\it clock}, which represents some unit of work in an ML algorithm, akin to iteration. Given $P$ workers, SSP assigns each worker a clock $c_p$ that is initially zero. Then, each worker repeats the following operations: (1) perform computation using shared parameters ${\bf x}$ stored in the PS, (2) make additive updates ${\bf u}$ to the PS, and (3) advance its own clock $c_p$ by 1. The SSP model limits fast workers' progress so that the clock difference between the fastest and slowest worker is $\le s$, $s$ being a specified staleness parameter. This is achieved via:

\paragraph{SSP Condition (informal):}
\vspace{-0.3cm}
Let $c$ be the clock of the fastest workers. They may not make further progress until all other workers' updates ${\bf u}_p$ that were made at clocks at or before $c-s-1$ become visible.


We present the formal condition in the next section. Crucially, there are multiple update communication strategies that can meet the SSP condition. We present {\it Eager SSP (ESSP)} as a class of implementations that eagerly propagate the updates to reduce empirical staleness beyond required by SSP. ESSP does not provide new guarantees besides warranted by SSP, but we show that by reducing the average staleness ESSP achieves faster convergence theoretically and empirically.


\begin{figure}[h]
\vspace{-0.4cm}
\centering
\includegraphics[width=0.7\linewidth]{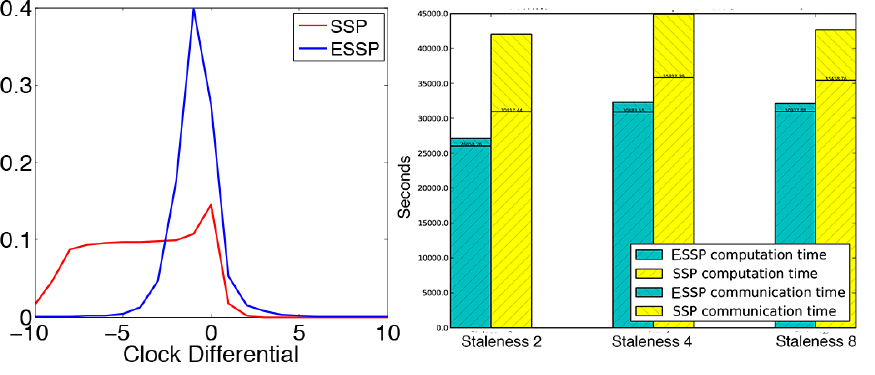}
\vspace{-0.4cm}
\caption{\small {\bf Left:} Empirical staleness distribution from matrix factorization. X-axis is (parameter age - local clock), y-axis is the normalized observation count. Note that on Bulk Synchronous Parallel (BSP) system such as Map-Reduce, the staleness is always $-1$. We use rank 100 for matrix factorization, and each clock is $1\%$ minibatch. The experiment is run on a 64 node cluster. {\bf Right:} Communication and Computation breakdown for LDA. The lower part of the bars are computation, and the upper part is communication.}
\label{fig:comm_comp}
\vspace{-0.4cm}
\end{figure}

How can we show that ESSP reduces the staleness of parameter reads? While it is difficult to {\it a priori} predict the behavior of complex software such as a PS, we can still empirically measure the staleness of parameter reads during PS algorithm execution, from which we can draw conclusions about the PS's behavior. Consider Figure~\ref{fig:comm_comp} (left), which shows the distribution of parameter stalenesses observed in matrix factorization implemented on SSP and ESSP.
Our measure of staleness is a ``clock differential": when a worker reads a parameter, that worker's clock could be be behind (or ahead) of other workers, and clock differential simply measures this clock difference. Under SSP, the distribution of clock differentials is nearly uniform, because SSP ``waits until the last minute" to update the local parameter cache.
On the other hand, ESSP frequently updates the local parameter caches via its eager communication, which
reduces the negative tail in clock differential distribution --- this improved staleness profile is ESSP's most salient advantage over SSP. In the sequel, we will show that better staleness profiles lead to faster ML algorithm convergence, by proving new, tighter convergence bounds based on average staleness and the staleness distributions (unlike the simpler worst-case bounds in~\cite{ho2013more}).

Our analyses and experiments show that ESSP combines the strengths of VAP and SSP: (1) ESSP achieves strong theoretical properties comparable to VAP; (2) ESSP can be efficiently implemented, with excellent empirical performance on two ML algorithms: matrix completion using SGD, and topic modeling using sampling. We also show that ESSP achieves higher throughput than SSP, thanks to system optimizations that exploit ESSP's aggressive scheduling.

\vspace{-0.3cm}
\section{Theoretical Analysis}
\label{sec:theory}



In this section, we theoretically analyze VAP and ESSP, and show how they affect ML algorithm convergence. For space reasons, all proofs are placed in the appendix. As explained earlier, we ground our analysis on ML algorithms in the stochastic gradient descent (SGD) family (due to its high popularity for Big Data), and
prove the convergence of SGD under VAP and ESSP. We now explain SGD in the context of a matrix completion problem.

\vspace{-0.2cm}
\subsection{SGD for Low Rank Matrix Factorization}

Matrix completion involves decomposing an $N\times M$ matrix $D$ into two low rank matrices $L\in \mathbb{R}^{N\times K}$ and $R\in \mathbb{R}^{K\times M}$ such that $LR\approx D$, where $K << \min\{M,N\}$ is a user-specified rank.
The problem is to predict those missing entries based on known entries $D_{obs}$ by solving the following $\ell_2$-penalized optimization problem:
\vspace{-0.3cm}
\begin{align*}
\min_{L,R} \sum_{(i,j)\in D_{obs}} ||D_{ij} - \sum_{k=1}^K L_{ik}R_{kj}||^2 + \lambda (||L||_F^2 + ||R||_F^2)
\end{align*}
where $||\cdot||_F$ is the Frobenius norm and $\lambda$ is the regularization parameter. The stochastic gradient updates for each observed entry $D_{ij}\in D_{obs}$ are
\vspace{-0.2cm}
\begin{align*}
L_{i*} &\leftarrow L_{i*} + \gamma(e_{ij}R_{*j}^{\top}-\lambda L_{i*})
\\
R_{*j}^{\top} &\leftarrow R_{*j} + \gamma(e_{ij}L_{i*}^{\top}-\lambda R_{*j})
\end{align*}
where $L_{i*}$, $R_{*j}$ are row and column of $L,R$ respectively, and $L_{i*}R_{*j}$ is the vector product. $e_{ij}=D_{ij}-L_{i*}R_{*j}$. We absorb constants into the step-size $\gamma$. Since $L,R$ are being updated by each gradient, we put them in the parameter server to allow all works access them and make additive updates. The data $D_{obs}$ are partitioned into worker nodes and stored locally. 




\vspace{-0.4cm}
\paragraph{VAP} We formally introduce the VAP computation model: given $P$ workers that produce updates at regular intervals which we call ``clocks'', let $\bu_{p,c}\in \mathbb{R}^n$ be the update from worker $p$ at clock $c$ applied to the system state $\bx\in \mathbb{R}^n$ via
$\bx \leftarrow \bx + \bu_{p,c}$. 
Consider the update sequence $\hbu_t$ that orders the updates based on the global time-stamp they are generated. We can define ``real-time sequence'' $\hbx_t$ as
\vspace{-0.3cm}
\begin{align*}
\hbx_t := \bx_0 + \sum_{t'=1}^t \hbu_{t'}
\end{align*}
assuming all workers start from the agreed-upon initial state $\bx_0$. (Note that $\hbx_t$ is different from the parameter server view as the updates from different workers can arrive the server in a different order due to network.) Let $\bbx_t$ be the noisy view some worker $w$ sees when generating update $\hbu_t$, i.e., $\hbu_t := G(\bbx_t)$ for some function $G$. The VAP condition guarantees
\begin{equation}
||\bbx_t - \hbx_t||_{\infty} \le v_t = \frac{v_0}{\sqrt{t}}
\end{equation}
where we require the value bound $v_t$ to shrink over time from the initial bound $v_0$. Notice that $\bbx_t - \hbx_t$ is exactly the updates {\it in transit} w.r.t. worker $w$. We make mild assumptions to avoid pathological cases.\footnote{To avoid pathological cases where a worker is delayed indefinitely, we assume that each worker's updates are finitely apart in sequence $\hbu_t$. In other words, all workers generate updates with sufficient frequency. For SGD, we further assume that each worker updates its step-sizes sufficiently often that the local step-size $\breve{\eta}_t = \frac{\eta}{\sqrt{t-r}}$ for some bounded drift $r\ge 0$ and thus $\breve{\eta}_t$ is close to the global step size schedule $\eta_t=\frac{\eta}{\sqrt{t}}$.}

\begin{theorem}
\label{thm:vap_expectation}
(SGD under VAP, convergence in expectation)
Given convex function $f(\bx)=\sum_{t=1}^T f_t(\bx)$ such that components $f_t$ are also convex. We search for minimizer $\bx^*$ via gradient descent on each component $\nabla f_t$ with step-size $\breve{\eta}_t$ close to $\eta_t = \frac{\eta}{\sqrt{t}}$ such that the update $\hbu_t = -\breve{\eta}_t \nabla f_t(\bbx_t)$ is computed on noisy view $\bbx_t$. The VAP bound follows the decreasing $v_t$ described above. Under suitable conditions ($f_t$ are $L$-Lipschitz and bounded diameter $D(x\Vert x^\prime) \le F^2$),
\begin{align*}
R[X] := \sum_{t=1}^T f_t(\bbx_t) - f(\bx^*) = \order{\sqrt{T}}
\end{align*}
and thus $\frac{R[X]}{T} \to 0$ as $T\rightarrow \infty$.
\end{theorem}

Theorem~\ref{thm:vap_expectation} implies that the worker's noisy VAP view $\bbx_t$ converges to the global optimum $\bx^*$, as measured by $f$, in expectation at the rate $\order{T^{-1/2}}$. The analysis is similar to~\cite{ho2013more}, but we use the real-time sequence $\hbx_t$ as our reference sequence and VAP condition instead of SSP. Loosely speaking, Theorem~\ref{thm:vap_expectation} shows that VAP execution is unbiased. We now present a new bound on the variance.

\begin{theorem}
(SGD under VAP, bounded variance)
Assuming $f(\bx)$, $\breve{\eta}_t$, and $v_t$ similar to theorem~\ref{thm:vap_expectation} above, and $f(\bx)$ has bounded and invertible Hessian, $\Omega^*$ defined at optimal point $\bx^*$. Let $\Var_t := \mE[\bbx_t^2]-\mE[\bbx_t]^2$ ($\Var_t$ is the sum of component-wise variance\footnote{$\Var_t = \sum_{i=1}^d \mE[\breve{x}_{ti}^2]- \mE[\breve{x}_{ti}]^2$}), and $\bbg_t = \nabla f_t(\bbx_t)$ is the gradient, then:
\begin{align}
\Var_{t+1} 
&= \Var_{t} -2cov(\hbx_t, \mE^{\Delta_t}[\bbg_t]) +\order{\delta_t}
\\
&+ \order{\breve{\eta}_t^2\rho_t^2} + \bigo^*_{\delta_t}
\label{eq:vap_var}
\end{align}
near the optima $\bx^{*}$. The covariance $cov(\bm{v}_1, \bm{v}_2) := \mE[\bm{v}_1^T \bm{v}_2] - \mE[\bm{v}_1^T] \mE[\bm{v}_2]$ uses inner product. $\delta_t = ||\bdelta_t||_{\infty}$ and $\bdelta_t = \bbx_t - \hbx_t$. $\rho_t = ||\bbx_t-\bx^*||$. $\Delta_t$ is a random variable capturing the randomness of update $\hbu_t = -\eta_t \bbg_t$ conditioned on $\hbx_t$ (see appendix).
\label{thm:vap_var}
\end{theorem}
$cov(\hbx_t, \mE^{\Delta_t}[\bbg_t]) \ge 0$ in general as the change in $\bx_t$ and average gradient $\mE^{\Delta_t}[\bbg_t]$ are of the same direction. Theorem~\ref{thm:vap_var} implies that under VAP the variance decreases in successive iterations for sufficiently small $\delta_t$, which can be controlled via VAP threshold $v_t$. However, as we argued in the previous section, the VAP condition requires the same amount of synchronization as strong consistency, which makes it of little practical benefit. This motivates our following analysis of the SSP model.

\vspace{-0.4cm}
\paragraph{SSP}
We return to the $(p,c)$ index. Under the SSP worker $p$ at clock $c$ only has access to a noisy view $\tilde{\bx}_{p,c}$ of the system state ($\tbx$ is different from the noisy view in VAP $\bbx$). Update $\bu_{p,c} = G(\tbx_{p,c})$ is computed on the noisy view $\tbx_{p,c}$ for some function $G()$. Assuming all workers start from the agreed-upon initial state $\bx_0$, the SSP condition is:

\vspace{-0.4cm}
\paragraph{SSP Bounded-Staleness (formal):} For a fixed staleness $s$, the noisy state $\tilde{\bx}_{p,c}$ is equal to
\begin{align*}
\tilde{\bx}_{p,c} &= \bx_0
+ \underset{\text{guaranteed pre-window updates}}{\underbrace{\left[ \sum_{c^\prime = 1}^{c-s-1} \sum_{p^\prime=1}^{P} \bu_{p^\prime,c^\prime} \right]}}
+ \underset{\text{best-effort in-window updates}}{\underbrace{\left[ \sum_{(p^\prime,c^\prime)\in \mathcal{S}_{p,c}} \bu_{p^\prime,c^\prime} \right]}},
\end{align*}
for some $\mathcal{S}_{p,c} \subseteq \mathcal{W}_{p,c} = \{1,...,P\} \times \{c-s,...,c+s-1\}$ which is some subset of updates in the $2s$ window issued by all $P$ workers during clock $c-s$ to $c+s-1$. The noisy view consists of (1) guaranteed pre-window updates for clock 1 to $c-s-1$, and (2) best-effort updates indexed by $\mathcal{S}_{p,c}$.\footnote{In contrast to~\cite{ho2013more}, we do not assume read-my-write.} We introduce a clock-major index $t$:
\begin{align*}
\tilde{\bx}_t &:= \tilde{\bx}_{(t\;\text{mod}\; P),\lfloor t/P \rfloor}
\qquad
&&\bu_t := \bu_{(t\;\text{mod}\; P),\lfloor t/P \rfloor}
\end{align*}
and analogously for $\mathcal{S}_t$ and $\mathcal{W}_t$. We can now define a reference sequence (distinct from $\hbx_t$ in VAP) which we informally refers to as the ``true'' sequence:
\begin{align}
\textstyle
\bx_t &= \bx_0 + \sum_{t^\prime=0}^{t} \bu_{t^\prime}
\label{eq:trueCond}
\end{align}
The sum loops over workers ($t$ mod $P$) and clocks $\lfloor t/P \rfloor$ . Notice that this sequence is unrelated to the server view.

\begin{theorem} (SGD under SSP, convergence in expectation~\cite{ho2013more}, Theorem 1)
Given convex function $f(\bx)=\sum_{t=1}^T f_t(\bx)$ with suitable conditions as in Theorem~\ref{thm:vap_expectation}, we use gradient descent with updates $\bu_t=-\eta_t \nabla f_t(\tbx_t)$ generated from noisy view $\tbx_t$ and $\eta_t = \frac{\eta}{\sqrt{t}}$. Then
\vspace{-0.4cm}
\begin{align*}
R[X] := \sum_{t=1}^T f_t(\tbx_t) - f(\bx^*) = \order{\sqrt{T}}
\end{align*}
and thus $\frac{R[X]}{T} \to 0$ as $T\rightarrow \infty$.
\label{thm:ssp_expectation}
\end{theorem}

Theorem~\ref{thm:ssp_expectation} is the SSP-counterpart of Theorem~\ref{thm:vap_expectation}. The analysis of Theorem~\ref{thm:ssp_expectation} only uses the worst-case SSP bounds. However, in practice many updates are much less stale than the SSP bound. In Fig~\ref{fig:comm_comp} (left) both implementations have small portion of updates with maximum staleness.

We now use moment statistics to further characterize the convergence. We begin by decomposing $\tbx_t$. Let $\bar{u}_t := \frac{1}{P(2s+1)} \sum_{t'\in \mathcal{W}_t} ||\bu_{t'}||_2$ be the average of $\ell_2$ of the updates. We can write the noisy view $\tilde{\bx}_t$ as
\begin{equation}
\tilde{\bx}_t = \bx_t + \bar{u}_t \bgamma_t
\label{eq:decomp}
\end{equation}
where $\bgamma_t \in \mathbb{R}^d$ is a vector of random variables whose randomness lies in the network communication. Note that the decomposition in eq.~\ref{eq:decomp} is always possible since $\bar{u}_t = 0$ iff $\bu_{t'} = \bm{0}$ for all updates $\bu_{t'}$ in the 2$s$ window. Using SSP we can bound $\bar{u}_t$ and $\bgamma_t$:
\begin{lemma}
$\bar{u}_t \le \frac{\eta}{\sqrt{t}} L$ and 
$\gamma_t := ||\bgamma_t||_2 \le P(2s+1)$.
\label{lemma:ubound}
\end{lemma}
Therefore $\mu_{\gamma} = \mE[\gamma_t]$ and $\sigma_{\gamma} = var(\gamma_t)$ are well-defined. We now provide an exponential tail-bound characterizing convergence in finite steps.

\begin{theorem} (SGD under SSP, convergence in probability)
Given convex function $f(\bx)=\sum_{t=1}^T f_t(\bx)$ such that components $f_t$ are also convex. We search for minimizer $\bx^*$ via gradient descent on each component $\nabla f_t$ under SSP with staleness $s$ and $P$ workers. Let $\bu_t:=-\eta_t \nabla_t f_t(\tilde{\bx}_t)$ with $\eta_t=\frac{\eta}{\sqrt{t}}$. Under suitable conditions ($f_t$ are $L$-Lipschitz and bounded divergence $D(x||x')\le F^2$), we have
\begin{align*}
P\left[\frac{R\left[X\right]}{T} - \frac{1}{\sqrt{T}} \left(\eta L^{2} + \frac{F^{2}}{\eta} + 2\eta L^2\mu_{\gamma} \right) \ge \tau\right] 
\\
\le \exp\left\{\frac{-T\tau^2}{2\bar{\eta}_T\sigma_{\gamma} + \frac{2}{3}\eta L^2(2s+1)P\tau}\right\}
\end{align*}
where $R[X] := \sum_{t=1}^T f_t(\tilde{x}_t) - f(x^*)$, and $\bar{\eta}_T = \frac{\eta^2 L^4 (\ln T + 1)}{T} = o(T)$.
\label{thm:sgd_tail}
\end{theorem}
This means that $\frac{R[X]}{T}$ converges to $O(T^{-1/2})$ in probability with an exponential tail-bound. Also note that the convergence is faster for smaller $\mu_{\gamma}$ and $\sigma_{\gamma}$.

We need a few mild assumptions on the staleness $\gamma_t$ in order to derive variance bound:
\begin{assumption}
$\gamma_{t}$ are i.i.d. random variable with well-defined mean $\mu_{\gamma}$ and variance $\sigma_{\gamma}$.
\end{assumption}
\begin{assumption}
$\gamma_t$ is independent of $\bx_t$ and $\bu_t$.
\end{assumption}
Assumption 1 is satisfied by Lemma~\ref{lemma:ubound}, while Assumption 2 is valid since $\gamma_t$ are only influenced by the computational load and network bandwidth at each machine, which are themselves independent of the actual values of the computation ($\bu_t$ and $\bx_t$). We now present an SSP variance bound.

\begin{theorem}
(SGD under SSP, decreasing variance) Given the setup in Theorem~\ref{thm:sgd_tail} and assumption 1-3. Further assume that $f(\bx)$ has bounded and invertible Hessian $\Omega^*$ at optimum $\bx^*$ and $\gamma_t$ is bounded. Let $\Var_t := \mE[\tbx_t^2]-\mE[\tbx_t]^2$, $\bm{g}_t=\nabla f_t(\tbx_t)$ then for $\tbx_t$ near the optima $\bx^{*}$ such that $\rho_t = ||\tbx_t-\bx^*||$ and $\xi_t = ||\bm{g}_t||-||\bm{g}_{t+1}||$ are small:
\begin{align}
\Var_{t+1}&= \Var_t - 2\eta_t cov(\bx_t, \mE^{\Delta_t}[\bm{g}_t]) + \order{\eta_t \xi_t}
\\
&+ \order{\eta_t^2 \rho_t^2} + \bigo^*_{\gamma_t}
\label{eq:ssp_var}
\end{align}
where the covariance $cov(\bm{v}_1, \bm{v}_2) := \mE[\bm{v}_1^T \bm{v}_2] - \mE[\bm{v}_1^T] \mE[\bm{v}_2]$ uses inner product. $\bigo^*_{\bgamma_t}$ represents high order ($\ge 5$th) terms involving $\gamma_t = ||\bgamma_t||_{\infty}$. $\Delta_t$ is a random variable capturing the randomness of update $\bu_t$ conditioned on $\bx_t$ (see appendix).
\label{thm:ssp_var}
\end{theorem}
As argued before, $cov(\bx_t, \mE^{\Delta_t}[\bm{g}_t]) \ge 0$ in general. Therefore the theorem implies that $\Var_t$ monotonically decreases over time when SGD is close to an optima.

\subsection{Comparison of VAP and ESSP}

From Theorem~\ref{thm:vap_var} and~\ref{thm:ssp_var} we see that both VAP and (E)SSP achieves decreasing variance. However, VAP convergence is much more sensitive to its tuning parameter (the VAP threshold) than (E)SSP, whose tuning parameter is the staleness $s$. This is evident from the $O(\delta_t)$ term in Eq.~\ref{eq:vap_var}, which is bounded by the VAP threshold. In contrast, (E)SSP's variance only involves staleness $\gamma_t$ in high order terms $\bigo^*_{\gamma_t}$ (Eq.~\ref{eq:ssp_var}), where $\gamma_t$ is bounded by staleness. This implies that staleness-induced variance vanishes quickly in (E)SSP. The main reason for (E)SSP's weak dependency on staleness because SGD's step-size already tunes the update magnitude approaching optimum, and thus for the same number of missing updates (such as the $2s$ window in SSP), their total magnitude is decreasing as well, which is conducive for lowering variance. VAP on the other hand, does not make use of the decreasing step-size and thus needs to directly rely on the VAP threshold, resulting in strong dependency on the threshold.

An intuitive analogy is that of postmen: VAP is like a postman who only ever delivers mail above a certain weight threshold $W$. (E)SSP is like a postman who delivers mail late, but no later than $T$ days. Intuitively, the (E)SSP postman is more reliable than the VAP postman due to his regularity. The only way for the VAP postman to be reliable, is to decrease the weight threshold $W\rightarrow 0$ --- this becomes important when the ML algorithm is approaching convergence, because the algorithm's updates become diminishingly small. However, there are two drawbacks to decreasing $W$: first, much like step-size tuning, it must be done at a carefully controlled rate --- this requires either specific knowledge about the ML problem, or a sophisticated, automatic scheme (that may also be domain-specific). Second, as $W$ decreases, VAP produces more communication, which increases the running time of the distributed algorithm.

In contrast to VAP, ESSP does not suffer as much from these drawbacks, because: (1) SSP has a weaker theoretical dependency on its staleness threshold (than VAP does on its value-bound threshold), thus it is usually unnecessary to decrease the staleness as the ML algorithm approaches convergence; this is evidenced by the SSP paper~\cite{ho2013more}, which achieved stable convergence even though they did not decrease staleness gradually during ML algorithm execution. (2) Because ESSP proactively pushes out fresh parameter values, the distribution of stale reads is usually close to zero-staleness, regardless of the actual staleness threshold used (see Figure \ref{fig:comm_comp}) --- hence, fine-grained tuning of the staleness threshold is rarely necessary under ESSP.

\section{ESSPTable: An efficient ESSP System}





Our theory suggests that the ESSP consistency model, with its aggressive parameter updates, should considerably outperform SSP. In order to verify this, we implemented ESSP inside a Parameter Server (PS), which we call ESSPTable.

\vspace{-0.3cm}
\paragraph{PS Interface:}
\vspace{-0.2cm}
Each physical machine runs one ESSPTable process with three types of threads: 1) computation threads; 2) communication threads and 3) server threads. Each computation thread is regarded as a distinct worker by the system. The computation threads execute application logic and access the global parameters stored in ESSPTable through a key-value store interface---read a table-row via \texttt{GET} and write via \texttt{INC}. 
Once a computation thread completes a clock tick, it notifies the system via \texttt{CLOCK}, which increments the worker's clock by 1. As required by the SSP consistency , a \texttt{READ} issued by a worker at clock $c$ is guaranteed to observe all updates generated in clock $[0, c - s - 1]$, where $s$ is the user-defined staleness threshold.

\vspace{-0.35cm}
\paragraph{Ensuring Consistency Guarantees:}
\vspace{-0.2cm}
The ESSPTable client library caches locally accessed parameters. In case the parameter is too large to fit into client machine's memory, cold parameters are evicted using an approximate Least-Recently-Used (LRU) policy. When a computation thread issues a \texttt{GET} request, it first checks the local cache
for the requested parameters. If the requested parameter is not found in local cache, a read request is sent to the server.

Each parameter in the client local cache is associated with a clock $c_{param}$; $c_{param} = x$ means that all updates from all workers generated before clock $x$ have already been applied to this parameter. 
$c_{param}$ is compared with the worker's clock $c_{worker}$. Only if $c_{param} > c_{worker} - s$, the requested parameter is returned to the worker.

\vspace{-0.3cm}
\paragraph{Communication Protocol:}
\vspace{-0.2cm}
The updates generated by computation threads are coalesced since they are commutative and associative. These updates are sent to the server at the end of each clock tick.

The server sends updated parameters to the client through call-backs. When a client request a table-row for the first time, it registers a call-back on the server. This is the only time the client makes read request to the server. Subsequently, when a server table's clock advances from getting the clock tick from all clients, it pushes out the table-rows to the respective registered clients. This differs from the SSPTable in~\cite{ho2013more} where the server passively sends out updates upon client's read request (which happens each time a client's local cache becomes too stale). The call-back mechanism exploits the fact that computation threads often revisit the same parameters in iterative-convergent algorithms, and thus the server can push out table-rows to registered clients without clients' explicit request. Our server-push model causes more eager communication and thus lower empirical staleness than SSPTable in~\cite{ho2013more} as shown in Fig.~\ref{fig:comm_comp} (left).

We empirically observed that the time needed to communicate the coalesced updates accumulated in one clock is usually less than the computation time. Thus computation threads usually observe parameters with staleness $1$ regardless of the user-specified staleness threshold $s$. That relieves the burden of staleness tuning. Also, since the server pushes out updated parameters to registered clients in batches, it reduces the overall latency from sending each parameter separately upon clients' requests (which is the case in SSPTable). This improvement is shown in our experiments.

\vspace{-0.3cm}
\section{Experiments}



We show that ESSP improves the speed and quality of convergence (versus SSP) for collapsed gibbs sampling in topic model and stochastic gradient descent (SGD) in matrix factorization. Furthermore, ESSP is robust against the staleness setting, relieving the user from worrying about an additional tuning parameter. The experimental setups are:

\begin{itemize}

\item {\bf ML Models and algorithms:} LDA topic modeling (using collapsed Gibbs sampling) and Matrix Factorization (stochastic gradient descent). Both algorithms are implemented using ESSPTable's interface. For LDA we use 50\% minibatch in each \texttt{Clock()} call, and we use log-likelihood as measure of training quality. For MF we use 1\% and 10\% minibatch in each \texttt{Clock()} and record the squared loss (instead of the $\ell_2$-penalized objective) for convenient comparison with GraphLab (see appendix). The step size for MF is chosen to be large while the algorithm still converges with staleness 0.

\item {\bf Datasets} Topic model: New York Times ($N=100m$ tokens, $V=100k$ vocabularies, and $K=100$ topics). Matrix factorization: Netflix dataset ($480k$ by $18k$ matrix with $100m$ nonzeros.) We use rank $K=100$.

\item {\bf Compute cluster} Matrix factorization experiments were run on 64 nodes, each with 2 cores and 16GB RAM, connected via 1Gbps ethernet. LDA experiments were run on 8 nodes, each with 64 cores and 128GB memory, connected via 1Gbps ethernet.

\end{itemize}

\begin{figure}[t]
\vspace{-0.4cm}
\centering
\includegraphics[width=1\linewidth]{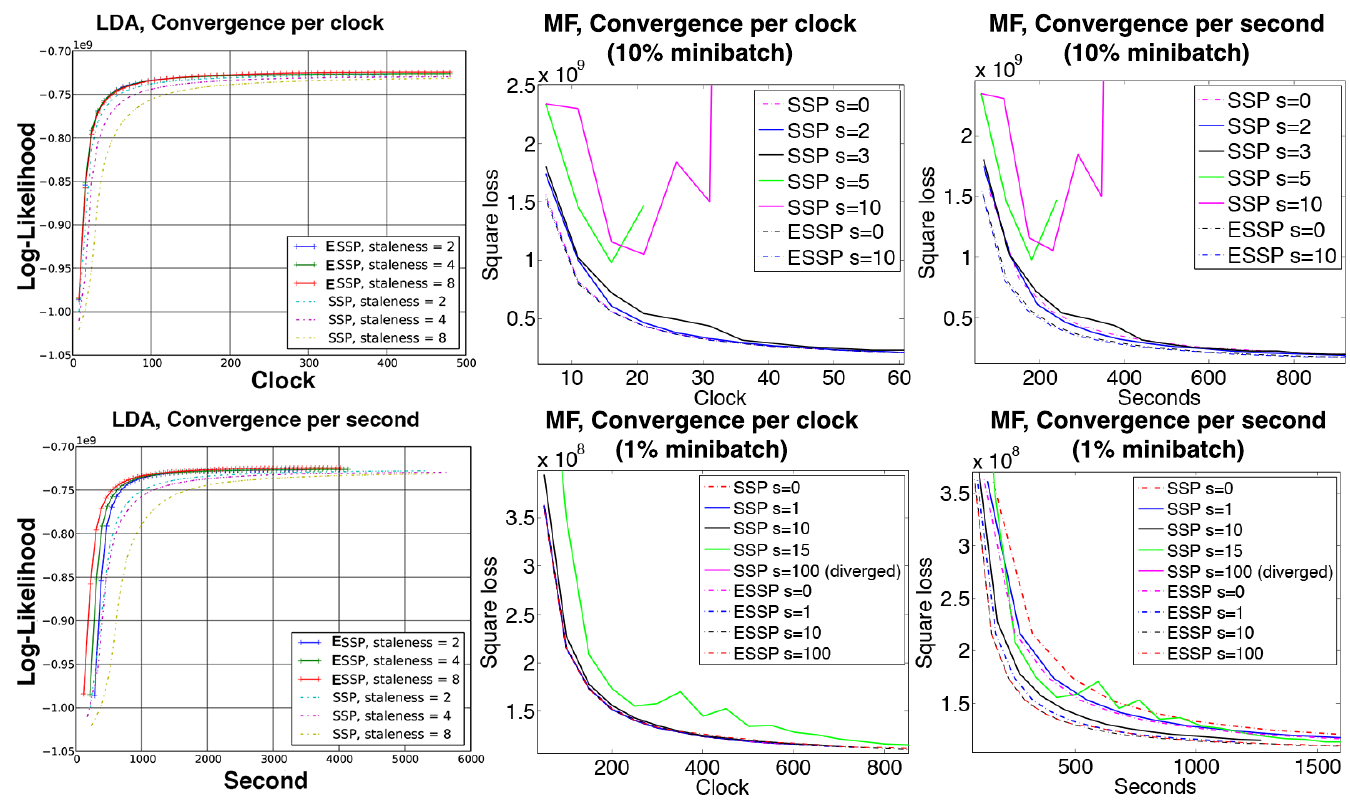}
\vspace{-0.6cm}
\caption{\small{\bf Experimental Results.} The convergence speed per iteration and per second for LDA and MF.}
\label{fig:panel}
\vspace{-0.4cm}
\end{figure}

{\bf Speed of Convergence:} Figure~\ref{fig:panel} shows the objective over iteration and time for LDA and matrix factorization. In both cases ESSP converges faster or comparable to SSP with respect to iteration and run time. The speed up over iteration is due to the reduced staleness as shown in the staleness profile (Figure~\ref{fig:comm_comp}, left). This is consistent with the fact that in SSP, computation making use of fresh data makes more progress~\cite{ho2013more}. Also it is worth pointing out that staleness helps SSP substantially but much less so for ESSP because ESSP is less susceptible to staleness.

{\bf Robustness to Staleness:} One important tuning knob in SGD-type of algorithms are the step size. Using step sizes that are too small leads to slow convergence, while step sizes that are too large cause divergence. The problem of stepsize tuning is aggravated in the distributed setting, where staleness could aggregate the updates in a non-deterministic manner, thus causing unpredictable performance (dependent on network congestion and the varying machine speeds). In the case of MF, SSP diverges under high staleness, as staleness effective increases the step size. However, ESSP is robust across all investigated staleness values due to the concentrated staleness profile (see Figure~\ref{fig:comm_comp}, left). Even when high SSP staleness does not produce divergence, the convergence is ``shaky'' due to the variance introduced by staleness. ESSP produces lower variance for all staleness settings, consistent with our theoretical analyses. This improvement largely reduced the need for user to tune the staleness parameter introduced in SSP.

{\bf System Opportunity.} In addition to faster convergence per iteration, ESSP provides opportunities for system to optimize the communication. By sending updates preemptively, ESSP not only reduces the staleness but also reduces the chance of client threads being blocked to wait for updates. In some sense ESSP is a more ``pipelined'' version of SSP. Figure~\ref{fig:comm_comp} (right) shows the breakdown of communication and compuation time for varying staleness. This contributes to the overall convergence per second in Figure~\ref{fig:panel}, where ESSP has larger margin of speed gain over SSP than the convergence per iteration.


\vspace{-0.3cm}
\section{Related Work and Discussion}


Existing software that is tailored towards {\it distributed} (rather than merely single-machine parallel), scalable ML can be roughly grouped into two categories: general-purpose, programmable libraries or frameworks such as GraphLab~\cite{graphlab10} and Parameter Servers (PSes)~\cite{ho2013more,li2013parameter}, or special-purpose solvers tailored to specific categories of ML applications: CCD++~\cite{yu2012scalable} for Matrix Factorization, Fugue~\cite{fugue} for constrained MF, Vowpal Wabbit for regression/classification problems via stochastic optimization~\cite{langford2007vowpal}, and Yahoo LDA as well as Google plda for topic modeling~\cite{plda}. 

The primary differences between the general-purpose frameworks (including this work) and the special-purpose solvers are (1) the former are user-programmable and can be extended to handle arbitrary ML applications, while the latter are non-programmable and restricted to predefined ML applications; (2) because the former must support arbitrary ML programs, their focus is on improving the ``systems" code (notably, communication and synchronization protocols for model state) to increase the efficiency of {\it all ML algorithms}, particularly through the careful design of consistency models (graph consistency in GraphLab, and iteration/value-bounded consistency in Parameter Servers) --- in contrast, the special-purpose systems combine both systems code improvements and {\it algorithmic} (i.e. mathematical) improvements tailor-made for their specific category of ML applications.

As a paper about general-purpose distributed ML, we focus on consistency models and systems code, and we deliberately use (relatively) simple algorithms for our benchmark applications, for two reasons: (1) to provide a fair comparison, we must match the code/algorithmic complexity of the benchmarks for other frameworks like GraphLab and SSP PS~\cite{ho2013more} (practically, this means our applications should use the same update equations); (2) a general-purpose ML framework should not depend on highly-specialized algorithmic techniques tailored only to specific ML categories. The crux is that general-purpose frameworks should {\it democratize distributed ML} in a way that special-purpose solvers cannot, by enabling those ML applications that have been under-served by the distributed ML research community to benefit from cluster computing. Since our benchmark applications are kept algorithmically simple, they are unlikely to beat the special-purpose solvers in running time --- but we note that many algorithmic techniques featured in those solvers can be applied to our benchmark applications, by dint of the general-purpose nature of PS programming.

In~\cite{muli}, the authors propose and implement a PS consistency model that has similar theoretical guarantees to the ideal VAP model presented herein. However, we note that their implementation does not strictly enforce the conditions of their consistency model. Their implementation implicitly assumes zero latency for transmission over network, while in a real cluster, there could be arbitrarily long network delay. In their system, reads do not wait for delayed updates, so a worker may compute
with highly inconsistent parameters in the case of congested network.

On the wider subject of Big Data, Hadoop~\cite{borthakur2007hadoop} and Spark~\cite{Zaharia_Spark10} are popular programming frameworks, which ML applications are sometimes developed on top of. To the best of our knowledge, there is no recent work showing that Hadoop or Spark have superior ML algorithm performance compared to frameworks designed for ML like GraphLab and PSes (let alone the special-purpose solvers mentioned earlier). The most salient difference is that Hadoop and Spark only feature strict consistency, and do not support flexible consistency models like graph- or bounded-consistency. On the positive side, Hadoop and Spark ensure program portability, reliability and fault tolerance at a level that GraphLab and PSes have yet to match.

{\small
\bibliography{main}

\begin{thebibliography}{10}

\bibitem{Ahmed:2012:SIL:2124295.2124312}
Amr Ahmed, Moahmed Aly, Joseph Gonzalez, Shravan Narayanamurthy, and
  Alexander~J. Smola.
\newblock Scalable inference in latent variable models.
\newblock In {\em WSDM}, pages 123--132, 2012.

\bibitem{borthakur2007hadoop}
Dhruba Borthakur.
\newblock The hadoop distributed file system: Architecture and design.
\newblock {\em Hadoop Project Website}, 11:21, 2007.

\bibitem{Bradley+al:icml11parlasso}
Joseph~K. Bradley, Aapo Kyrola, Danny Bickson, and Carlos Guestrin.
\newblock Parallel coordinate descent for l1-regularized loss minimization.
\newblock In {\em International Conference on Machine Learning (ICML 2011)},
  June 2011.

\bibitem{adam}
Trishul Chilimbi, Yutaka Suzue, Johnson Apacible, and Karthik Kalyanaraman.
\newblock Project adam: Building an efficient and scalable deep learning
  training system.
\newblock In {\em 11th USENIX Symposium on Operating Systems Design and
  Implementation (OSDI 14)}, pages 571--582, Broomfield, CO, October 2014.
  USENIX Association.

\bibitem{cipar2013}
James Cipar, Qirong Ho, Jin~Kyu Kim, Seunghak Lee, Gregory~R. Ganger, Garth
  Gibson, Kimberly Keeton, and Eric Xing.
\newblock Solving the straggler problem with bounded staleness.
\newblock In {\em HotOS '13}. Usenix, 2013.

\bibitem{cui_atc_14}
Henggang Cui, James Cipar, Qirong Ho, Jin~Kyu Kim, Seunghak Lee, Abhimanu
  Kumar, Jinliang Wei, Wei Dai, Gregory~R. Ganger, Phillip~B. Gibbons, Garth~A.
  Gibson, and Eric~P. Xing.
\newblock Exploiting bounded staleness to speed up big data analytics.
\newblock In {\em 2014 USENIX Annual Technical Conference (USENIX ATC 14)},
  pages 37--48, Philadelphia, PA, June 2014. USENIX Association.

\bibitem{dean2012large}
J~Dean, G~Corrado, R~Monga, K~Chen, M~Devin, Q~Le, M~Mao, M~Ranzato, A~Senior,
  P~Tucker, K~Yang, and A~Ng.
\newblock Large scale distributed deep networks.
\newblock In {\em NIPS 2012}, 2012.

\bibitem{ho2013more}
Q.~Ho, J.~Cipar, H.~Cui, J.-K. Kim, S.~Lee, P.~B. Gibbons, G.~Gibson, G.~R.
  Ganger, and E.~P. Xing.
\newblock More effective distributed ml via a stale synchronous parallel
  parameter server.
\newblock In {\em NIPS}, 2013.

\bibitem{fugue}
Abhimanu Kumar, Alex Beutel, Qirong Ho, and Eric~P. Xing.
\newblock Fugue: Slow-worker-agnostic distributed learning for big models on
  big data.
\newblock Technical report.

\bibitem{langford2007vowpal}
J~Langford, L~Li, and A~Strehl.
\newblock Vowpal wabbit online learning project, 2007.

\bibitem{lee2014strads}
Seunghak Lee, Jin~Kyu Kim, Xun Zheng, Qirong Ho, Garth~A Gibson, and Eric~P
  Xing.
\newblock Primitives for dynamic big model parallelism.
\newblock In {\em Advances in Neural Information Processing Systems (NIPS)},
  2014.

\bibitem{muli_osdi14}
Mu~Li, David~G. Andersen, Jun~Woo Park, Alexander~J. Smola, Amr Ahmed, Vanja
  Josifovski, James Long, Eugene~J. Shekita, and Bor-Yiing Su.
\newblock Scaling distributed machine learning with the parameter server.
\newblock In {\em Operating Systems Design and Implementation (OSDI)}, 2014.

\bibitem{muli}
Mu~Li, Li~Zhou~Zichao Yang, Aaron Li~Fei Xia, David~G. Andersen, and Alexander
  Smola.
\newblock Parameter server for distributed machine learning.
\newblock {\em NIPS workshop}, 2013.

\bibitem{li2013parameter}
Mu~Li, Li~Zhou, Zichao Yang, Aaron Li, Fei Xia, Dave Andersen, and Alex Smola.
\newblock Parameter server for distributed machine learning, big learning
  workshop.
\newblock In {\em NIPS}, 2013.

\bibitem{graphlab10}
Yucheng Low, Joseph Gonzalez, Aapo Kyrola, Danny Bickson, Carlos Guestrin, and
  Joseph~M. Hellerstein.
\newblock Graphlab: A new parallel framework for machine learning.
\newblock In {\em Conference on Uncertainty in Artificial Intelligence (UAI)},
  Catalina Island, California, July 2010.

\bibitem{niu2011hogwild}
Feng Niu, Benjamin Recht, Christopher R{\'e}, and Stephen~J Wright.
\newblock Hogwild!: A lock-free approach to parallelizing stochastic gradient
  descent.
\newblock In {\em NIPS}, 2011.

\bibitem{plda}
Yi~Wang, Hongjie Bai, Matt Stanton, Wen-Yen Chen, and Edward~Y. Chang.
\newblock Plda: Parallel latent dirichlet allocation for large-scale
  applications.
\newblock In {\em Proceedings of the 5th International Conference on
  Algorithmic Aspects in Information and Management}, AAIM '09, pages 301--314,
  Berlin, Heidelberg, 2009. Springer-Verlag.

\bibitem{yu2012scalable}
Hsiang-Fu Yu, Cho-Jui Hsieh, Si~Si, and Inderjit~S Dhillon.
\newblock Scalable coordinate descent approaches to parallel matrix
  factorization for recommender systems.
\newblock In {\em ICDM}, pages 765--774, 2012.

\bibitem{Zaharia_Spark10}
Matei Zaharia, N.~M.~Mosharaf Chowdhury, Michael Franklin, Scott Shenker, and
  Ion Stoica.
\newblock Spark: Cluster computing with working sets.
\newblock Technical Report UCB/EECS-2010-53, EECS Department, University of
  California, Berkeley, May 2010.

\end{thebibliography}
\bibliographystyle{plain}
}

\section{Appendix}

\paragraph{Theorem 1}
(SGD under VAP, convergence in expectation)
Given convex function $f(\bx)=\sum_{t=1}^T f_t(\bx)$ such that components $f_t$ are also convex. We search for minimizer $\bx^*$ via gradient descent on each component $\nabla f_t$ with step-size $\breve{\eta}_t$ close to $\eta_t = \frac{\eta}{\sqrt{t}}$ such that the update $\hbu_t = -\breve{\eta}_t \nabla f_t(\bbx_t)$ is computed on noisy view $\bbx_t$. The VAP bound follows the decreasing $v_t$ described above. Under suitable conditions ($f_t$ are $L$-Lipschitz and bounded diameter $D(x\Vert x^\prime) \le F^2$),
\begin{align*}
R[X] := \sum_{t=1}^T f_t(\bbx_t) - f(\bx^*) = \order{\sqrt{T}}
\end{align*}
and thus $\frac{R[X]}{T} \to 0$ as $T\rightarrow \infty$.

\begin{proof}

We will use real-time sequence $\hbx_t$ defined by
\begin{align*}
\hbx_t := \bx_0 + \sum_{t'=1}^t \hbu_{t'}
\end{align*}

\begin{align*}
R[X] &= \sum_{t=1}^T f_t(\bbx_t) - f(\bx^*)
\\
&\le \sum_{t=1}^T \langle \nabla f_t(\bbx_t), \bbx_t - \bx^*\rangle &&\text{($f_t$ are convex)}
\\
&=\sum_{t=1}^T \langle \bbg_t, \bbx_t - \bx^* \rangle
\end{align*}

where $\bbg_t := \nabla f_t(\bbx_t)$. From Lemma A.1 below we have

\begin{align*}
R[X] &\le \sum_{t=1}^T \frac{1}{2} \breve{\eta}_t ||\bbg_t||^2 + \frac{D(\bx^*|| \hbx_t) - D(\bx^*|| \hbx_{t+1})} {\breve{\eta}_t} + \langle \bbx_t - \hbx_t, \bbg_t \rangle
\end{align*}

We now bound each term:

\begin{align*}
\sum_{t=1}^T \frac{1}{2} \breve{\eta}_t ||\bbg_t||^2 &\le \sum_{t=1}^T \frac{1}{2} \breve{\eta}_t L^2 &&\text{(Lipschitz assumption)}
\\
&= \sum_{t=r+1}^T \frac{1}{2} \frac{\eta}{\sqrt{t-r}} L^2 + const &&\text{($r>0$ is the finite clock drift in VAP)}
\\
&= \frac{1}{2}\eta L^2\sum_{t=r+1}^T  \frac{1}{\sqrt{t-r}} + const
\\
&\le\frac{1}{2}\eta L^2\int_{t=r+1}^T  \frac{1}{\sqrt{t-r}} dt + const
\\
&\le\frac{1}{2}\eta L^2 (\sqrt{T-r} - 1) + const
\\
&= \order{\sqrt{T}}
\end{align*}
where the clock drift comes from the fact that $\breve{\eta}_t$ is not exactly $\eta_t=\frac{\eta}{\sqrt{t}}$ in VAP.

\begin{align*}
\sum_{t=1}^T \frac{D(\bx^*|| \hbx_t) - D(\bx^*|| \hbx_{t+1})} {\breve{\eta}_t} &= \frac{D(\bx^*|| \hbx_1)}{\breve{\eta}_1} - \frac{D(\bx^*|| \hbx_{T+1})}{\breve{\eta}_T} + \sum_{t=2}^T \left[ D(\bx^*||\hbx_t) \left( \frac{1}{\breve{\eta}_t} - \frac{1}{\breve{\eta}_{t-1}} \right) \right]
\\
&\le \frac{F^2}{\eta} + 0 + \frac{F^2}{\eta}\sum_{t=2}^T \left[ \sqrt{t-k} - \sqrt{t-r} \right] \quad \text{(clock drift)}
\\
&\le\frac{F^2}{\eta} + \frac{F^2}{\eta}\int_{t=\max(k,r)}^T \left(\sqrt{t-k}-\sqrt{t-r}\right) dt + const
\\
&=\frac{F^2}{\eta} + \frac{F^2}{\eta} \left[ (t-k)^{3/2} - (t-r)^{3/2} \right]_{max(k,r)}^T + const
\\
&=\frac{F^2}{\eta} + \frac{F^2}{\eta} \left[(T-k)^{3/2} - (T-r)^{3/2} \right] + const
\\
&=\frac{F^2}{\eta} + \frac{F^2}{\eta}\left[ \left(T^{\frac{3}{2}} + \frac{3}{2}kT^{\frac{1}{2}} + \order{\sqrt{T}}\right) \right.
\\
& \left.- \left(T^{\frac{3}{2}} + \frac{3}{2}rT^{\frac{1}{2}} + \order{\sqrt{T}}\right) \right] + const
\quad\text{(binomial expansion)}
\\
&= \order{\sqrt{T}}
\end{align*}

\begin{align*}
\sum_{t=1}^T \langle \bbx_t - \hbx_t, \bbg_t \rangle &\le \sum_{t=1}^T||\bbx_t - \hbx_t||_2 ||\bbg_t||_2
\\
&\le \sum_{t=1}^T \sqrt{d} v_t L \quad\text{(using eq.(2) from main text)}
\\
&= \sqrt{d}L \sum_{t=1}^T  \frac{v_0}{\sqrt{t}}
\\
&= \sqrt{d}L v_0\sqrt{T} = \order{\sqrt{T}}
\end{align*}

Together, we have $R[X]\le \order{\sqrt{T}}$ as desired.

\end{proof}

\paragraph{Lemma A.1 }
For $\bx^*, \bbx_t \in X$, and $X = \mathbb{R}^d$,
\begin{align*}
\langle \bbg_t, \bbx_t - \bx^* \rangle = \frac{1}{2} \breve{\eta}_t ||\bbg_t||^2 + \frac{D(\bx^*|| \hbx_t) - D(\bx^*|| \hbx_{t+1})} {\breve{\eta}_t} + \langle \bbx_t - \hbx_t, \bbg_t \rangle
\end{align*}
where $D(x||x'):=\frac{1}{2}||x-x'||^2$.

\begin{proof}
\begin{align*}
D(\bx^*|| \hbx_t) - D(\bx^*|| \hbx_{t+1}) &= \frac{1}{2}||\bx^*-\hbx_t + \hbx_t - \hbx_{t+1}||^2 - \frac{1}{2}||\bx^* - \hbx_t||^2
\\
&= \frac{1}{2}||\bx^*-\hbx_t + \breve{\eta}_t \bbg_t||^2 - \frac{1}{2}||\bx^* - \hbx_t||^2
\\
&=\frac{1}{2}\breve{\eta}_t ||\bbg_t||^2 - \breve{\eta}_t \langle \hbx_t-\bx^*, \bbg_t \rangle
\end{align*}

Divide both sides by $\breve{\eta}_t$ gets the desired answer.

\end{proof}

\noindent{\bf Lemma 4 }{\it
$\bar{u}_t \le \frac{\eta}{\sqrt{t}} L$ and $\gamma_t := ||\bgamma_t||_2 \le P(2s+1)$.
}

\begin{proof}
$||\bu_t||_2 = ||-\eta_t \nabla f_t||_2 \le \frac{\eta}{\sqrt{t}} L$ since $f$ is $L$-Lipschitz. Therefore $\bar{u}_t = \frac{1}{P(2s+1)} \sum_{t'\in \mathcal{W}_t} ||\bu_{t'}||_2 \le \frac{\eta}{\sqrt{t}}L$ since $|\mathcal{W}_t| \le P(2s+1)$.

If $\bar{u}_t = 0$, then $\bgamma_t=\bm{0}$ and the lemma holds trivially. For $\bar{u}_t > 0$. $\bgamma_t = \frac{1}{\bar{u}_t}(\tilde{\bx}_t - \bx_t) = \frac{1}{\bar{u}_t} \sum_{t'\in \mathcal{S}_t} \bu_{t'}$. Thus $||\bgamma_t||_2 = \frac{1}{\bar{u}_t} ||\sum_{t'\in \mathcal{S}_t} \bu_{t'}||_2 \le \frac{1}{\bar{u}_t} \sum_{t'\in \mathcal{S}_t} ||\bu_{t'}||_2 \le \frac{1}{\bar{u}_t} \sum_{t'\in \mathcal{W}_t} ||\bu_{t'}||_2 = P(2s+1)$.
\end{proof}

\noindent{\bf Theorem 5 }
{\it (SGD under SSP, convergence in probability) Given convex function $f(\bx)=\sum_{t=1}^T f_t(\bx)$ such that components $f_t$ are also convex. We search for minimizer $\bx^*$ via gradient descent on each component $\nabla f_t$ under SSP with staleness $s$ and $P$ workers. Let $\bu_t:=-\eta_t \nabla_t f_t(\tilde{\bx}_t)$ with $\eta_t=\frac{\eta}{\sqrt{t}}$. Under suitable conditions ($f_t$ are $L$-Lipschitz and bounded divergence $D(x||x')\le F^2$), we have
\begin{align*}
P\left[\frac{R\left[X\right]}{T} - \frac{1}{\sqrt{T}} \left(\eta L^{2} + \frac{F^{2}}{\eta} + 2\eta L^2\mu_{\gamma} \right) \ge \tau\right] 
\le \exp\left\{\frac{-T\tau^2}{2\bar{\eta}_T\sigma_{\gamma} + \frac{2}{3}\eta L^2(2s+1)P\tau}\right\}
\end{align*}
where $R[X] := \sum_{t=1}^T f_t(\tilde{x}_t) - f(x^*)$, and $\bar{\eta}_T = \frac{\eta^2 L^4 (\ln T + 1)}{T} = o(T)$.
}

\begin{proof}

From lemma A.1, substitute $\bbx_t$ with $\tilde{x}_t$ we have

\begin{eqnarray*}
R\left[X\right] & \le &\sum_{t=1}^{T}\left\langle \tilde{g}_{t},\tilde{x}_{t}-x^{*}\right\rangle
\\
& = & \sum_{t=1}^{T}\frac{1}{2}\eta_{t}\left\Vert \tilde{g}_{t}\right\Vert ^{2}+\frac{D\left(x^{*}\Vert x_{t}\right)-D\left(x^{*}\Vert x_{t+1}\right)}{\eta_{t}} + \left\langle \tilde{x}_{t}-x_{t},\tilde{g}_{t}\right\rangle
\\
&\le&  \eta L^{2}\sqrt{T} + \frac{F^{2}}{\eta}\sqrt{T} + \sum_{t=1}^{T} \left\langle \bar{u}_t\bgamma_t,\tilde{g}_{t}\right\rangle
\\
&\le&  \eta L^{2}\sqrt{T} + \frac{F^{2}}{\eta}\sqrt{T} + \sum_{t=1}^{T} \frac{\eta}{\sqrt{t}} L^{2}\gamma_t 
\end{eqnarray*}
Where the last step uses the fact
\begin{align*}
\left\langle \bar{u}_t \bgamma_t ,\tilde{g}_{t}\right\rangle
&\le \bar{u}_t|| \bgamma_t||_2 ||\tilde{g}_t||_2
\\
&\le \gamma_t \frac{\eta}{\sqrt{t}} L^2  && \text{(Lemma 4)}
\end{align*}
Dividing $T$ on both sides,
\begin{eqnarray}
\frac{R\left[X\right]}{T} - \frac{\eta L^{2}}{\sqrt{T}} - \frac{F^{2}}{\eta\sqrt{T}} &\le& \frac{\sum_{t=1}^{T} \frac{\eta}{\sqrt{t}} L^{2}\gamma_t}{T}
\label{eq:risk_bound1}
\end{eqnarray}
Let $a_t:=\frac{\eta}{\sqrt{t}} L^2 (\gamma_t - \mu_{\gamma})$. Notice that $a_t$ zero-mean, and $|a_t|\le \eta L^2\max_t(\gamma_t) \le \eta L^2(2s+1)P$. Also, $\frac{1}{T}\sum_{t=1}^T var(a_t) = \frac{1}{T}\sum_{t=1}^T\frac{\eta^2}{t} L^4 \sigma_{\gamma} < \frac{\eta^2 L^4 \sigma_{\gamma}}{T}(\ln T + 1) = \bar{\eta}_T \sigma_{\gamma}$ where $\bar{\eta}_T=\frac{\eta^2 L^4 (\ln T + 1)}{T}$. Bernstein's inequality gives, for $\tau > 0$,
\begin{equation}
P\left(\frac{\sum_{t=1}^{T} \frac{\eta}{\sqrt{t}} L^{2}\gamma_t -\frac{\eta}{\sqrt{t}} L^2 \mu_{\gamma}}{T}  \ge \tau\right)  \le \exp\left\{\frac{-T\tau^2}{2\bar{\eta}_T \sigma_{\gamma} + \frac{2}{3}\eta L^2(2s+1)P\tau}\right\}
\label{eq:bernstein}
\end{equation}

Note the following identity:
\begin{equation}
\sum_{i=a}^b\frac{1}{\sqrt{i}} \le 2\sqrt{b-a+1}
\end{equation}
Thus
\begin{equation}
\frac{1}{T}\sum_{t=1}^T \frac{\eta}{\sqrt{t}} L^2 \mu_{\gamma} \le \frac{2\eta L^2 \mu_{\gamma}}{\sqrt{T}}
\label{eq:mean_bound}
\end{equation}

Plugging eq.~\ref{eq:risk_bound1} and~\ref{eq:mean_bound} to eq.~\ref{eq:bernstein}, we have
\begin{equation*}
P\left[\frac{R\left[X\right]}{T} - \frac{1}{\sqrt{T}} \left(\eta L^{2} + \frac{F^{2}}{\eta} + 2\eta L^2\mu_{\gamma} \right) \ge \tau\right]  \le \exp\left\{\frac{-T\tau^2}{2\bar{\eta}_T\sigma_{\gamma} + \frac{2}{3}\eta L^2(2s+1)P\tau}\right\}
\end{equation*}

\end{proof}

We need the following Lemma to prove Theorem 2 and 6.

\paragraph{Lemma A.2}
Let $\bm{\Omega^*}$ be the hessian of the
loss at optimum $\bx^*$, then
\begin{equation*}
\bm{g}_t := \nabla f(\tbx_{t}) = (\tbx_{t}-\bx^*)\bm{\Omega^*} + \order{\rho_{t}^2}
\end{equation*}
for $\tbx_{t}$ close to the optimum such that $\order{\rho_{t}} = \order{||\tbx_{t}-\bx^*||}$ is small. Here $\bm{\Omega^*} = \left.\nabla^2 f(\bx) \right |_{\bx=\bx^*}$ is the Hessian at the optimum

\begin{proof}
Using Taylor's theorem and expanding around $\bx^{*}$,
\begin{align*}
f(\tbx_{t})
&=f(\bx^*)+(\tbx_{t}-\bx^*)^T \left.\nabla f(\bx)\right |_{\bx=\bx^*}
\\
&\;\;\;+\frac{1}{2}(\tbx_{t}-\bx^*)^T \bm{\Omega^*}(\tbx_{t}-\bx^*)
+\order{||\tbx_{t}-\bx^{*}||^3}
\\
&=f(\bx^*)+\frac{1}{2}(\tbx_{t}-\bx^*)^T \bm{\Omega^*}(\tbx_{t}-\bx^*)
+\order{||\tbx_{t}-\bx^*||^3}
\end{align*}
where the last step uses $\nabla f(\bx)=0$ at $\bx^*$. Taking gradient w.r.t. $\tbx_{t}$,
\begin{align*}
\nabla f(\tbx_{t})
&= (\tbx_{t} - \bx^*)^T \bm{\Omega^*} + \order{||\tbx_{t}-\bx^*||^2}
\\
&= (\tbx_{t} - \bx^*)^T \bm{\Omega^*} + \order{\rho_{t}^2}
\end{align*}
\end{proof}

\paragraph{Theorem 6}
(SGD under SSP, decreasing variance) Given the setup in Theorem 5 and assumption 1-3. Further assume that $f(\bx)$ has bounded and invertible Hessian $\Omega^*$ at optimum $\bx^*$ and $\gamma_t$ is bounded. Let $\Var_t := \mE[\tbx_t^2]-\mE[\tbx_t]^2$, $\bm{g}_t=\nabla f_t(\tbx_t)$ then for $\tbx_t$ near the optima $\bx^{*}$ such that $\rho_t = ||\tbx_t-\bx^*||$ and $\xi_t = ||\bm{g}_t||-||\bm{g}_{t+1}||$ are small:
\begin{align*}
\Var_{t+1}&= \Var_t - 2\eta_t cov(\bx_t, \mE^{\Delta_t}[\bm{g}_t]) + \order{\eta_t \xi_t}
\\
&+ \order{\eta_t^2 \rho_t^2} + \bigo^*_{\gamma_t}
\end{align*}
where the covariance $cov(\bm{v}_1, \bm{v}_2) := \mE[\bm{v}_1^T \bm{v}_2] - \mE[\bm{v}_1^T] \mE[\bm{v}_2]$ uses inner product. $\bigo^*_{\bgamma_t}$ represents high order ($\ge 5$th) terms involving $\gamma_t = ||\bgamma_t||_{\infty}$. $\Delta_t$ is a random variable capturing the randomness of update $\bu_t$ conditioned on $\bx_t$.

\begin{proof}
We write eq. 3 from the main text as $\tilde{\bx_t} = \bx_t + \bdelta_t$ with $\bdelta_t = \bar{u}_t\bgamma_t$. Conditioned on $\bx_t$, we have
\begin{equation}
p(\tbx_t|\bx_t) d\tbx_t = p(V_t(\bdelta_t,\bx_t)) dV_t
\label{eq:volSpace}
\end{equation}
where $V_t$ is a random variable representing the state of $\bdelta_t$ conditioned on $\bx_t$. We can express $\mE^{\tbx_t}[f(\tbx_t)]$ in terms of $\mE^{\bx_t}$ for any function $f()$ of $\tbx_t$:
\begin{align}
\mE^{\tbx_t}[f(\tbx_t)] &= \int_{\tbx_t} f(\tbx_t) p(\tbx_t) d\tbx_t
\nonumber
\\
&=\int_{\tbx_t}\int_{\bx_t} f(\tbx_t) p(\tbx_t|\bx_t) p(\bx_t) d\bx_t d\tbx_t && \text{(using eq.~\ref{eq:volSpace})}\nonumber \\
&=\int_{\bx_t}\int_{V_t} f(\tbx_t) p(V_t(\bdelta_t,\bx_t)) dV_t d\bx_t \nonumber
\\
&=\mE^{\bx_t}\left[\mE^{V_t} [f(\tbx_t) ]\right]
\label{eq:ex2_re}
\end{align}

Similarly, we have
\begin{equation}
\mE^{\tbx_{t+1}}[f(\tbx_{t+1})]=\mE^{\bx_{t+1}}\left[\mE^{V_{t+1}}[f(\tbx_{t+1})]\right]
\end{equation}

In the same vein, we introduce random variable $\Delta$, conditioned on $\bx_t$:
\begin{equation}
p(\bx_{t+1}|\bx_t) d\bx_{t+1} = p(\Delta_t(\bu_t,\bx_t)) d\Delta_t
\label{eq:volSpaceU}
\end{equation}
since $\bx_{t+1}=\bx_t + \bu_t$ (eq. 2 in the main text). Here $\Delta$ is a random variable representing the state of $\bu_t$ conditioned on $\bx_t$. Analogous to eq.~\ref{eq:ex2_re}, we have
\begin{align}
\mE^{\bx_{t+1}}[f(\bx_{t+1})] 
&=\mE^{\bx_t}[\mE^{\Delta_t}[f(\bx_{t+1})]]
\label{eq:ex_plus1_re}
\end{align}
for some function $f()$ of $\bx_{t+1}$. There are a few facts we will use throughout:
\begin{align}
\mE^{\bx_t}\left[h(\bx_t, \bar{u}_t) \mE^{V_t}[\bgamma_t] \right]
&= \mE^{\bx_t}[h(\bx_t, \bar{u}_t)] \mE^{V_t}[\bgamma_t] && \text{(since $\bgamma_t \bot \bx_t,\bar{u}_t$)}
\\
\mE^{\bx_t}\left[ \mE^{\Delta_t}[\bx_t^T g(\bu_t)] \right] &= \mE^{\bx_t}\left[\bx_t^T \mE^{\Delta_t}[g(\bu_t)] \right] &&\text{($\Delta_t$ conditioned on $\bx_t$)}
\label{eq:push_delta}
\\
\mE^{\Delta_t}[\bar{u}_{t+1}] &= \bar{u}_{t+1}
\label{eq:delta_u}
\end{align}
where $h(\bx_t,\bar{u}_t)$ is some function of $\bx_t$ and $\bar{u}_t$, and similarly for $g()$. Eq.~\ref{eq:delta_u} follows from $\bar{u}_{t+1}$ being an average over the randomness represented by $\Delta_t$. We can now expand $\Var_t$:
\begin{align}
\Var_t &= \mE^{\tbx_t}[\tbx_t^2] - (\mE^{\tbx_t}[\tbx_t])^2 \nonumber
\\
&= \mE^{\bx_{t}}[\mE^{V_t}[\tbx_t^2]]  -  (\mE^{\bx_{t}}[\mE^{V_t}[\tbx_{t}]])^2 &&\text{(using eq.~\ref{eq:ex2_re})} \nonumber
\\
&=\mE^{\bx_{t}}[\mE^{V_{t}}[\bx_{t}^2 + \bdelta_{t}^2 + 2\bx_{t}^T\bdelta_{t}]] - 
(\mE^{\bx_{t}}[\mE^{V_{t}}[\bx_{t}+\bdelta_{t}]])^2 \label{eq:commonVar}
\end{align}
We expand each term:
\begin{align*}
&\mE^{\bx_{t}}[\mE^{V_{t}}[\bx_{t}^2 + \bdelta_{t}^2 + 2\bx_{t}^T\bdelta_{t}]]
\\
&= \mE^{\bx_{t}}[\bx_{t}^2 + \mE^{V_{t}}[\bdelta_{t}^2] + 2\bx_{t}^T\mE^{V_{t}}[\bdelta_{t}]]
\\
&= \mE^{\bx_{t}}[\bx_{t}^2] + \mE^{\bx_{t}}[\bar{u}_t^2\mE^{V_{t}}[\bgamma_{t}^2]] + 2\mE^{\bx_{t}}[\bx_{t}^T\bar{u}_t\mE^{V_{t}}[\bgamma_{t}]]
\\
&= \mE^{\bx_{t}}[\bx_{t}^2] + \mE^{\bx_{t}}[\bar{u}_t^2]\mE^{V_{t}}[\bgamma_{t}^2] + 2\mE^{\bx_{t}}[\bx_{t}^T\bar{u}_t]\mE^{V_{t}}[\bgamma_{t}]
\end{align*}
\begin{align*}
&(\mE^{\bx_{t}}[\mE^{V_{t}}[\bx_{t}+\bdelta_{t}]])^2
\\
&= (\mE^{\bx_{t}}[\bx_{t}+\mE^{V_{t}}[\bdelta_{t}]])^2
\\
&= (\mE^{\bx_{t}}[\bx_{t}+\bar{u}_t \mE^{V_{t}}[\bgamma_{t}]])^2
\\
&= (\mE^{\bx_{t}}[\bx_{t}]+\mE^{\bx_{t}}[\bar{u}_t]\mE^{V_{t}}[\bgamma_{t}]])^2
\\
&= \mE^{\bx_{t}}[\bx_{t}]^2 + \mE^{\bx_{t}}[\bar{u}_t]^2\mE^{V_{t}}[\bgamma_{t}]^2 + 2\mE^{\bx_{t}}[\bx_{t}^T]\mE^{\bx_{t}}[\bar{u}_t]\mE^{V_{t}}[\bgamma_{t}]
\end{align*}
Therefore
\begin{equation}
\begin{split}
\Var_t
&=\mE^{\bx_{t}}[\bx_{t}^2] + \mE^{\bx_{t}}[\bar{u}_t^2]\mE^{V_{t}}[\bgamma_{t}^2] + 2\mE^{\bx_{t}}[\bx_{t}^T\bar{u}_t]\mE^{V_{t}}[\bgamma_{t}]
\\
&- \mE^{\bx_{t}}[\bx_{t}]^2 - \mE^{\bx_{t}}[\bar{u}_t]^2\mE^{V_{t}}[\bgamma_{t}]^2 - 2\mE^{\bx_{t}}[\bx_{t}^T]\mE^{\bx_{t}}[\bar{u}_t]\mE^{V_{t}}[\bgamma_{t}]
\end{split}
\label{eq:varT}
\end{equation}

Following similar procedures, we can write $\Var_{t+1}$ as
\begin{equation}
\begin{split}
\Var_{t+1}
&=\mE^{\bx_{t+1}}[\bx_{t+1}^2] + \mE^{\bx_{t+1}}[\bar{u}_{t+1}^2]\mE^{V_{t+1}}[\bgamma_{t+1}^2]
\\
& +2\mE^{\bx_{t+1}}[\bx_{t+1}^T\bar{u}_{t+1}]\mE^{V_{t+1}}[\bgamma_{t+1}]
\\
& -\mE^{\bx_{t+1}}[\bx_{t+1}]^2 - \mE^{\bx_{t+1}}[\bar{u}_{t+1}]^2\mE^{V_{t+1}}[\bgamma_{t+1}]^2
\\
&- 2\mE^{\bx_{t+1}}[\bx_{t+1}^T]\mE^{\bx_{t+1}}[\bar{u}_{t+1}]\mE^{V_{t+1}}[\bgamma_{t+1}]
\end{split}
\label{eq:varT}
\end{equation}

We tackle each term separately:
\begin{align*}
\mE^{\bx_{t+1}}[\bx_{t+1}^2]
&= \mE^{\bx_t}\left[\mE^{\Delta_t}[(\bx_t + \bu_t)^2]\right] &&\text{(using eq.~\ref{eq:ex_plus1_re}, 2 main text)}
\\
&= \mE^{\bx_t}[\bx_t^2] + \mE^{\bx_t}\left[\mE^{\Delta_t}[\bu_t^2]\right] + 2\mE^{\bx_t}\left[ \bx_t^T \mE^{\Delta_t}[\bu_t] \right] &&\text{(using eq.~\ref{eq:push_delta})}
\end{align*}


\begin{align*}
&2\mE^{\bx_{t+1}}[\bx_{t+1}^T\bar{u}_{t+1}]\mE^{V_{t+1}}[\bgamma_{t+1}]
\\
&= 2\mE^{\bx_t}\left[\mE^{\Delta_t}[(\bx_t + \bu_t)^T \bar{u}_{t+1}]\right] \mE^{V_{t+1}}[\bgamma_{t+1}] &&\text{(using eq.~\ref{eq:ex_plus1_re}, 2 main text)}
\\
&= 2\mE^{\bx_t}\left[\mE^{\Delta_t}[\bx_t^T \bar{u}_{t+1}]\right] \mE^{V_{t+1}}[\bgamma_{t+1}]
\\
&+2\mE^{\bx_t}\left[\mE^{\Delta_t}[\bu_t^T \bar{u}_{t+1}]\right] \mE^{V_{t+1}}[\bgamma_{t+1}]
\\
&= 2\mE^{\bx_t}\left[\bx_t^T \bar{u}_{t+1}\right] \mE^{V_{t+1}}[\bgamma_{t+1}] &&\text{(using eq.~\ref{eq:push_delta} and \ref{eq:delta_u})}
\\
&+2\mE^{\bx_t}\left[\mE^{\Delta_t}[\bu_t^T \bar{u}_{t+1}]\right] \mE^{V_{t+1}}[\bgamma_{t+1}]
\end{align*}

\begin{align*}
-\mE^{\bx_{t+1}}[\bx_{t+1}]^2
&= -\mE^{\bx_t}\left[\mE^{\Delta_t}[\bx_t + \bu_t]\right]^2
\\
&= -\mE^{\bx_t}[\bx_t]^2 - \mE^{\bx_t}\left[\mE^{\Delta_t}[\bu_t]\right]^2 - 2\mE^{\bx_t}[\bx_t^T]\mE^{x_t}\left[ \mE^{\Delta_t}[\bu_t] \right]
\end{align*}

\begin{align*}
&-2\mE^{\bx_{t+1}}[\bx_{t+1}^T]\mE^{\bx_{t+1}}[\bar{u}_{t+1}]\mE^{V_{t+1}}[\bgamma_{t+1}]
\\
&= -2\mE^{\bx_{t}}\left[\mE^{\Delta_t}[(\bx_{t}+\bu_t)^T]\right]\mE^{\bx_{t}}\left[\mE^{\Delta_t}[\bar{u}_{t+1}]\right]\mE^{V_{t+1}}[\bgamma_{t+1}]
\\
&= -2\mE^{\bx_{t}}\left[\mE^{\Delta_t}[\bu_t^T]\right] \mE^{\bx_{t}}[\bar{u}_{t+1}]\mE^{V_{t+1}}[\bgamma_{t+1}]
-2\mE^{\bx_{t}}[\bx_t^T] \mE^{\bx_{t}}[\bar{u}_{t+1}]\mE^{V_{t+1}}[\bgamma_{t+1}]
\end{align*}

Assuming stationarity for $\bgamma_t$, and thus $\bar{\bgamma} := \mE^{V_t}[\bgamma_t] = \mE^{V_{t+1}}[\bgamma_{t+1}]$, we have
\begin{align*}
\Var_{t+1} - \Var_t
&= 2\left\{ \mE^{\bx_t}\left[ \bx_t^T \mE^{\Delta_t}[\bu_t] \right] - \mE^{\bx_t}[\bx_t^T] \mE^{\bx_t}\left[ \mE^{\Delta_t}[\bu_t] \right] \right\}
\\
&- 2\left\{ \mE^{\bx_t}[\bx_t^T(\bar{u}_t -\bar{u}_{t+1}) \bar{\bgamma}]- \mE^{\bx_t}[\bx_t^T] \mE^{\bx_t}[(\bar{u}_t - \bar{u}_{t+1})\bar{\bgamma}] \right\}
\\
&+\left\{ \mE^{\bx_t}\left[ \mE^{\Delta_t}[\bu_t^2] \right] + \mE^{\bx_{t+1}}[\bar{u}_{t+1}^2]\mE^{V_{t+1}}[\bgamma_{t+1}^2]- \mE^{\bx_t}\left[ \mE^{\Delta_t}[\bu_t] \right]^2 
\right.
\\
&-\mE^{\bx_t}[\bar{u}_{t+1}]^2 \bar{\bgamma}^2 - \mE^{\bx_t}[\bar{u}_t^2]\mE^{V_t}[\bgamma_t^2] +\mE^{\bx_t}[\bar{u}_t]^2\mE^{V_t}[\bgamma_t^2]
\\
&\left. +2\mE^{\bx_t}\left[ \mE^{\Delta_t}[\bu_t^T \bar{u}_{t+1}] \right] \bar{\bgamma}
-2\mE^{\bx_t}\left[\mE^{\Delta_t}[\bu_t^T] \right] \mE^{\bx_t}[\bar{u}_{t+1}]\bar{\bgamma}
\right\}
\\
&=2cov(\bx_t, \mE^{\Delta_t}[\bu_t]) + \order{\eta_t \xi_t} + \order{\eta_t^2 \rho_t^2} +\bigo^{*}
\end{align*}
where $\xi_t = ||\bm{g}_t|| - ||\bm{g}_{t+1}||$ and $\bigo^{*}$ are higher order terms. In the last step we use the fact that $||\bm{g}_t|| = \order{\rho_t}$ (lemma A.2) and thus $||\bu_t|| = \eta_t ||\nabla f(\bx_t)||$ and $\bar{u}_t$ are both $\order{\eta_t \rho_t}$. Notice that $cov(\bm{v}_1, \bm{v}_2) := \mE[\bm{v}_1^T \bm{v}_2] - \mE[\bm{v}_1^T] \mE[\bm{v}_2]$ uses inner product. Thus,
\begin{equation}
\Var_{t+1}= \Var_t - 2\eta_t cov(\bx_t, \mE^{\Delta_t}[\bm{g}_t]) + \order{\eta_t \xi_t} + \order{\eta_t^2 \rho_t^2} + \bigo^{*}
\end{equation}

\end{proof}

\paragraph{Theorem 2}
(SGD under VAP, bounded variance)
Assuming $f(\bx)$, $\breve{\eta}_t$, and $v_t$ similar to theorem 1, and $f(\bx)$ has bounded and invertible Hessian, $\Omega^*$ defined at optimal point $\bx^*$. Let $\Var_t := \mE[\bbx_t^2]-\mE[\bbx_t]^2$ ($\Var_t$ is the sum of component-wise variance\footnote{$\Var_t = \sum_{i=1}^d \mE[\breve{x}_{ti}^2]- \mE[\breve{x}_{ti}]^2$}), and $\bbg_t = \nabla f_t(\bbx_t)$ is the gradient, then:
\begin{align*}
\Var_{t+1} 
&= \Var_{t} -2cov(\hbx_t, \mE^{\Delta_t}[\bbg_t]) +\order{\delta_t}
+ \order{\breve{\eta}_t^2\rho_t^2} + \bigo^*_{\delta_t}
\end{align*}
near the optima $\bx^{*}$. The covariance $cov(\bm{v}_1, \bm{v}_2) := \mE[\bm{v}_1^T \bm{v}_2] - \mE[\bm{v}_1^T] \mE[\bm{v}_2]$ uses inner product. $\delta_t = ||\bdelta_t||_{\infty}$ and $\bdelta_t = \bbx_t - \hbx_t$. $\rho_t = ||\bbx_t-\bx^*||$. $\Delta_t$ is a random variable capturing the randomness of update $\hbu_t = -\eta_t \bbg_t$ conditioned on $\hbx_t$.

\begin{proof}
The proof is similar to the proof of Theorem 6. Starting off with $\bbx_t = \hbx_t + \bdelta_t$, we define $V_t$, $\Delta_t$ analogously. We have

\begin{align*}
\Var_t &= \mE^{\hbx_t}[\hbx_t^2] + \mE^{\hbx_t}[\mE^{V_t}[\bdelta_t^2]] + 2\mE^{\hbx_t}[\hbx_t^T \mE^{V_t}[\bdelta_t]]
\\&- \mE^{\hbx_t}[\hbx_t]^2 - \mE^{\hbx_t}[\mE^{V_t}[\bdelta_t^2]] - 2\mE^{\hbx_t}[\hbx_t]\mE^{\hbx_t^T}[\mE^{V_t}[\bdelta_t]]
\end{align*}

Similar algebra as in Theorem 6 leads to


\begin{align*}
\Var_{t+1}-\Var_t
&= 2cov(\hbx_t, \mE^{\Delta_t}[\hbu_t]) + 2cov(\hbx_t, \mE^{V_t}[\bdelta_t] - \mE^{\Delta_t}[\mE^{V_{t+1}}[\bdelta_{t+1}]])
\\
&+ \order{\delta_t^2} + \order{\breve{\eta}_t^2\rho_t^2} + \order{\breve{\eta}_t\delta_t} + \bigo^*
\\
&= -2cov(\hbx_t, \mE^{\Delta_t}[\bbg_t]) +\order{\delta_t}
+ \order{\breve{\eta}_t^2\rho_t^2} + \bigo^*_{\delta_t}
\end{align*}
where $\delta_t = ||\bdelta_t||_{\infty}$. This is the desired result in the theorem statement.
\end{proof}


\end{document}